\documentclass{article}

\usepackage{iclr2027_conference,times}
\usepackage{amsmath,amssymb}
\usepackage{booktabs}
\usepackage{array}
\usepackage{multirow}
\usepackage{graphicx}
\usepackage{microtype}
\usepackage{hyperref}
\usepackage{url}
\usepackage{xcolor}

\title{Bypass Observation: A Conceptual Design of a Non-Intrusive Layer-Wise Semantic Extraction Architecture}

\author{
Haibin Tong \\
Henan University \\
Kaifeng, China \\
\texttt{LBTHB@163.com}
\And
Jiang Yu\thanks{Corresponding author.} \\
National University of Defense Technology \\
Changsha, China \\
\texttt{yujiang@nudt.edu.cn}
}

\iclrfinalcopy

\begin{document}

\maketitle

\fancyhead{}
\renewcommand{\headrulewidth}{0pt}

\begin{abstract}
The reasoning process of large language models unfolds in a high-dimensional hidden-vector space, while humans can observe only the final output. This mismatch constitutes a central obstacle to interpretability and safety auditing. This paper systematically consolidates and refines a \emph{Bypass Observation} architecture. Without changing the vector transmission between layers of the backbone model, a read-only observation head is attached in parallel to selected inference layers. In its simplest form, the head is an LM-head-style or calibrated vocabulary readout that exposes intermediate predictive distributions; in an extended form, a separately trained semantic translator may convert these readouts or hidden states into compact human-readable summaries. In all cases, the observation output is never fed back into the backbone computation. We make four main contributions. First, for the full-vocabulary readout setting, we derive a closed-form approximation of the computational overhead: the additional cost is primarily governed by $V/(12d)$ and ranges from roughly 30\% to 240\% for representative models. We further discuss sparse observation, low-rank factorization, reduced semantic vocabularies, top-$k$ operators, and key-position sampling as practical cost-reduction strategies. Second, we argue that bypass observation moves a model from a black box toward a semi-transparent system. Its read-only nature prevents the observation output from causally influencing the backbone and separates the observation channel from conventional self-generated post-hoc explanations; however, the resulting readout remains only a partial view of the underlying hidden state, creating a potential \emph{white-box illusion}. Third, we distinguish the causal status of \emph{bypass chain-of-thought} from conventional chain-of-thought: the former is an open-loop readout, whereas the latter is part of a closed-loop computation. This distinction changes how the two signals should be interpreted and optimized in reinforcement learning. Fourth, we analyze the particular value of bypass observation for looped or recurrent-depth Transformers, where iteration-wise decoding can turn otherwise opaque dynamic computation into an observable convergence trajectory and may support halting criteria and safety thresholds. All conclusions in the present paper are conceptual or analytical and remain to be validated through systematic experiments.
\end{abstract}

\section{Introduction}

During inference in Transformer-based large language models, information is propagated through the residual stream as high-dimensional hidden vectors and is projected into a vocabulary distribution only at the output layer. Humans can directly observe the final decoded output, but the computations performed across tens or even hundreds of intermediate layers remain difficult to interpret. This black-box property creates at least three problems. First, it is difficult to answer what the model is computing at intermediate stages. Second, potentially deceptive internal states, the formation of hallucinations, or abnormal activations may remain invisible until the final output is produced. Third, reinforcement learning based only on terminal outcomes provides sparse credit assignment and may be vulnerable to linguistic reward hacking.

Existing interpretability methods can be grouped into several broad categories, each with structural limitations:
\begin{itemize}
    \item \textbf{Probes and linear decoding.} Probe-based methods~\citep{alain2016,hewitt2019} demonstrate that syntactic or semantic information can be decoded from hidden states, but such results primarily establish correlation rather than causation and often require offline training.
    \item \textbf{Lens methods.} The Logit Lens~\citep{nostalgebraist2020} directly applies the unembedding matrix to intermediate hidden states to inspect the evolution of predictions. The Tuned Lens~\citep{belrose2023} improves early-layer readability through layer-wise affine calibration. These methods, however, are typically used as diagnostic tools rather than as persistent architectural components.
    \item \textbf{Sparse autoencoders and mechanistic interpretability.} Sparse autoencoders (SAEs) extract interpretable features from superposed representations~\citep{cunningham2024,templeton2024}, while attribution-graph approaches organize features into computational structures~\citep{anthropic2025}. These methods can be powerful, but feature labeling often requires additional interpretation and the resulting analysis is usually sparse or retrospective.
    \item \textbf{Chain-of-thought and faithfulness studies.} Explicit intermediate reasoning can improve performance~\citep{wei2022}, but later work has shown that verbalized rationales are not always faithful descriptions of the computation that produced the answer~\citep{lanham2023,turpin2023}.
\end{itemize}

Bypass Observation can be viewed as an architectural, systematic, and controllable extension of the Logit Lens idea. Instead of treating layer-wise decoding as an occasional diagnostic operation, the model is equipped with a persistent observation pathway that can be deployed, evaluated, and potentially incorporated into training or auditing workflows. Its defining property is \textbf{non-intrusiveness}: decoded observations are exposed only to an external observer and are never inserted into the attention context of later layers.

\begin{figure}[t]
    \centering
    \includegraphics[width=\linewidth]{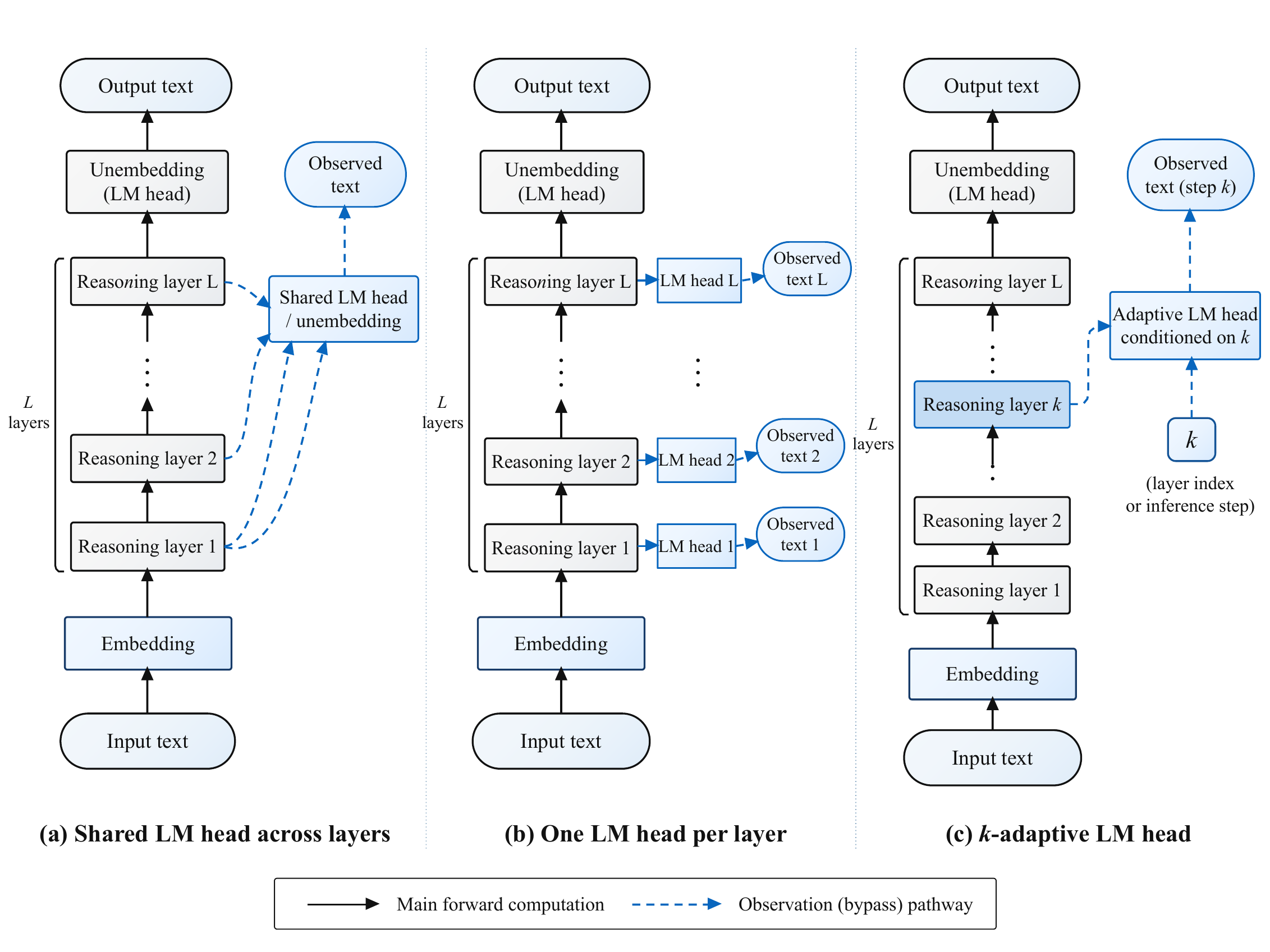}
    \caption{
    \textbf{Three architectural variants of Bypass Observation.}
    (a) A shared LM head is reused across multiple reasoning layers, providing a parameter-efficient common readout.
    (b) Each reasoning layer is equipped with its own LM head, allowing layer-specific calibration at increased parameter cost.
    (c) A $k$-adaptive LM head conditions its readout on the layer index or inference step, enabling one observer to adapt across depth or recurrent computation.
    Solid black arrows denote the main forward computation, while dashed blue arrows denote the read-only observation (bypass) pathway; the displayed ``observed text'' should be interpreted as a schematic readout that may correspond to top-$k$ vocabulary predictions or, with an additional semantic translator, a textual summary.
    In all variants, observation outputs are not fed back into the backbone.
    }
    \label{fig:bypass-framework}
\end{figure}

Figure~\ref{fig:bypass-framework} summarizes three compatible instantiations of the observation pathway: 
(a) a single LM head shared across multiple layers, 
(b) separate layer-specific LM heads, and 
(c) a $k$-adaptive LM head conditioned on the layer index or inference step.
These variants differ in parameter sharing and calibration flexibility, but all preserve the same read-only separation from the backbone computation.

\paragraph{Property 1 (Non-interference).}
Bypass Observation is a read-only operation attached to the backbone computation graph. Therefore, assuming deterministic execution and identical numerical kernels, attaching the observation module does not alter the functional input to any subsequent backbone layer and does not change the model's final output.

Existing empirical evidence from lens-based methods supports the premise that intermediate hidden states contain decodable predictive information. In particular, the Tuned Lens demonstrates that layer-specific calibration can substantially improve intermediate predictive readout. Bypass Observation adopts this evidence as motivation, while parameter sharing across layers remains an additional architectural choice rather than a property established by Tuned Lens itself. More generally, the value and the limitations of the proposed architecture must be stated together: Bypass Observation may provide a useful projection of an actual internal computation, but it does not provide a complete description of that computation.

\section{Formalization and Basic Axioms}

Consider a Transformer backbone with $L$ layers, hidden dimension $d$, and vocabulary size $V$. Let the hidden state at layer $\ell$ be
$\mathbf{h}_{\ell}\in\mathbb{R}^{d}$. Standard inference follows
\begin{equation}
    \mathbf{h}_{\ell+1}=F_{\ell}(\mathbf{h}_{\ell}).
\end{equation}

For selected layers, Bypass Observation attaches an observation function $O_{\ell}$ in parallel:
\begin{equation}
    y_{\ell}=\mathrm{Decode}\!\left(O_{\ell}(\mathbf{h}_{\ell})\right).
\end{equation}
Here, $O_{\ell}$ may denote either a lightweight vocabulary readout (for example, an LM head or a calibrated lens) or a learned semantic translator. The closed-form cost analysis in Section~3 applies specifically to the full-vocabulary readout case; a more expressive sequence decoder would introduce additional computation not captured by the $dV$ term.

We define the architecture through three axioms:
\begin{itemize}
    \item \textbf{(A1) Non-interference.}
    For every layer, the function $F_{\ell}$ and its inputs contain no observation output $y_k$ for $k\leq\ell$. The backbone output is therefore unchanged by the existence of the observation path. A1 is a defining property rather than an engineering preference: once $y_{\ell}$ is fed back into the backbone context, the system becomes a different architecture and the theoretical properties discussed here no longer apply.

    \item \textbf{(A2) Parameter sharing.}
    Observation heads may share parameters across layers, optionally conditioned on a layer index or a recurrent time-step embedding. This sharing is a design choice of Bypass Observation rather than a requirement inherited from Tuned Lens, whose calibration is typically layer-specific. For a shared full-vocabulary readout, the additional parameter count can be $O(dV)$ or lower, compared with a rough backbone scale of $O(12Ld^2)$.

    \item \textbf{(A3) Conditional readout.}
    The observation head may be conditioned on layer index or recurrent time step to account for representation drift and role changes across depth or iterations.
\end{itemize}

\section{Computational Overhead}

\subsection{Closed-Form Approximation}

We count multiply-accumulate operations (MACs) per generated token and per Transformer layer. One MAC corresponds to approximately two FLOPs.

Relevant hyperparameters include hidden dimension $d$, the number of query heads $h$, head dimension $d_{\mathrm{head}}$, the number of key/value heads $h_{\mathrm{kv}}$, feed-forward dimension $d_{\mathrm{ff}}$, and the feed-forward activation structure. The approximation below is independent of the total number of layers $L$.

\paragraph{Attention projections.}
For standard multi-head attention (MHA) with
$h\,d_{\mathrm{head}}=d$, the Q, K, V, and output projections each have effective shape $d\times d$, producing approximately
\begin{equation}
    C_{\mathrm{attn,proj}}\approx 4d^2
\end{equation}
MACs per layer.

With grouped-query attention (GQA) or multi-query attention (MQA), where $h_{\mathrm{kv}}<h$, the K and V projections are reduced to
$d\times(h_{\mathrm{kv}}d_{\mathrm{head}})$. The corresponding projection cost becomes
\begin{equation}
    C_{\mathrm{attn,proj}}
    =
    \left[
        2+2\left(\frac{h_{\mathrm{kv}}}{h}\right)
    \right]d^2 .
\end{equation}
During autoregressive decoding with a KV cache, the attention-score computation $QK^\top$ and multiplication by $V$ add approximately $2Sd$ MACs for context length $S$. Relative to a $12d^2$ layer estimate, this term has ratio $S/(6d)$; for $d=4096$, it becomes comparable only at context lengths on the order of $2.4\times10^4$ tokens. This estimate does not describe full-sequence prefill, whose attention term has the familiar quadratic dependence on sequence length.

\paragraph{Feed-forward network.}
A conventional GELU MLP with $d_{\mathrm{ff}}=4d$ contains an up-projection and a down-projection, giving
\begin{equation}
    C_{\mathrm{FFN}}
    =
    2dd_{\mathrm{ff}}
    =
    8d^2.
\end{equation}
Modern architectures such as Llama and Qwen commonly use a SwiGLU-style gated feed-forward block with three matrices. With a width near $(8/3)d$, its cost is also approximately
\begin{equation}
    3d\left(\frac{8}{3}d\right)=8d^2.
\end{equation}

\paragraph{Standard approximation.}
Under either MHA + GELU-$4d$ or MHA + SwiGLU-$(8/3)d$, the backbone cost per layer is therefore approximately
\begin{equation}
    C_{\mathrm{layer}}
    \approx
    4d^2+8d^2
    =
    12d^2
\end{equation}
MACs.

A full vocabulary observation head performs a $d\times V$ matrix-vector product at each observed layer, requiring approximately $dV$ MACs. If every layer is observed, the relative overhead is therefore
\begin{equation}
    \rho
    \approx
    \frac{dV}{12d^2}
    =
    \frac{V}{12d}.
\end{equation}

For arbitrary GQA and feed-forward widths, a more general expression is
\begin{align}
    C_{\mathrm{layer}}
    &=
    \left[
        2+2\left(\frac{h_{\mathrm{kv}}}{h}\right)
    \right]d^2
    +
    n_{\mathrm{ff}}dd_{\mathrm{ff}},\\
    \rho
    &=
    \frac{V}
    {
        \left[
            2+2\left(h_{\mathrm{kv}}/h\right)
        \right]d
        +
        n_{\mathrm{ff}}d_{\mathrm{ff}}
    },
\end{align}
where $n_{\mathrm{ff}}=2$ for a two-matrix GELU MLP and $n_{\mathrm{ff}}=3$ for a three-matrix SwiGLU block.

Layer normalization, softmax, residual additions, and biases involve primarily $O(d)$ element-wise operations and are ignored in this first-order estimate. The embedding lookup and final LM head are also omitted from the per-layer comparison. The approximation is related to the common $N\approx12Ld^2$ scaling heuristic discussed in compute analyses such as Chinchilla~\citep{hoffmann2022}. Architectures using more aggressively compressed attention representations may deviate further from this simplified formula.

\subsection{Representative Model Estimates}

Table~\ref{tab:cost} summarizes analytical estimates for representative model configurations.

\begin{table}[t]
\caption{Analytical estimates of the per-layer overhead of full-vocabulary bypass decoding. Values are computed from representative model configurations under the accounting assumptions described in Section~3.1.}
\label{tab:cost}
\centering
\resizebox{\linewidth}{!}{
\begin{tabular}{lrrrrrclrrrr}
\toprule
Model &
$d$ &
$V$ &
$h$ &
$d_{\mathrm{head}}$ &
$h_{\mathrm{kv}}$ &
Attention &
$d_{\mathrm{ff}}$ &
QKVO/$d^2$ &
FFN/$d^2$ &
Total/$d^2$ &
$\rho$ \\
\midrule
GPT-3 175B  & 12288 & 50257  & 96 & 128 & 96 & MHA & 49152 (4.0$d$)  & 4.00 & 8.00  & 12.00 & $\approx +34\%$ \\
Llama-2 7B  & 4096  & 32000  & 32 & 128 & 32 & MHA & 11008 (2.69$d$) & 4.00 & 8.06  & 12.06 & $\approx +65\%$ \\
Llama-2 70B & 8192  & 32000  & 64 & 128 & 8  & GQA & 28672 (3.5$d$)  & 2.25 & 10.50 & 12.75 & $\approx +31\%$ \\
Llama-3 8B  & 4096  & 128256 & 32 & 128 & 8  & GQA & 14336 (3.5$d$)  & 2.50 & 10.50 & 13.00 & $\approx +241\%$ \\
Qwen2.5 7B  & 3584  & 152064 & 28 & 128 & 4  & GQA & 18944 (5.29$d$) & 2.29 & 15.86 & 18.14 & $\approx +234\%$ \\
\bottomrule
\end{tabular}}
\end{table}

\paragraph{Observation 3.1.}
The overhead is primarily controlled by the ratio between vocabulary size $V$ and the effective per-layer width. Under the standard $12d^2$ approximation it reduces to $V/(12d)$. GQA decreases attention cost, while wider feed-forward blocks increase it, so realistic layer costs may vary from roughly $12d^2$ to $18d^2$ in the examples considered here. Consequently, the simple $V/(12d)$ estimate can deviate substantially from a model-specific calculation. The key reason for the large overhead is that a full LM head is itself a large $d\times V$ matrix. For modern models with vocabularies above 100k tokens, applying such a head at every layer can cost as much as, or more than, a complete Transformer layer.

\subsection{Engineering Cost and Cost-Reduction Strategies}

FLOPs are not the only bottleneck. During prefill, decoding every position at every layer can create a logits tensor with shape
\[
    \texttt{batch}\times\texttt{sequence}\times L\times V,
\]
which can become a more severe memory and bandwidth problem than the raw arithmetic count. Several reduction strategies are therefore important:

\begin{enumerate}
    \item \textbf{Sparse layer observation.}
    Observe one out of every $s$ layers, reducing the additional arithmetic cost approximately to $\rho/s$, where $s$ is the observation stride.

    \item \textbf{Low-rank readout.}
    Approximate the observation matrix by $W\approx BC$. If the intermediate rank is $r$, the cost scales approximately with $r/d+r/V$ relative to the full matrix. For example, a rank-64 factorization can drastically reduce the arithmetic cost for $d$ in the thousands and $V$ above 100k.

    \item \textbf{Reduced semantic vocabulary.}
    Equip the observation pathway with a smaller semantic vocabulary, for example on the order of 8k concepts or tokens, instead of the full model vocabulary.

    \item \textbf{Top-$k$ operators and key-position sampling.}
    Avoid materializing the complete $V$-dimensional distribution at every position. Observation can instead be restricted to sentence boundaries, answer spans, or other high-value positions.

    \item \textbf{Asynchronous deployment.}
    Enable Bypass Observation during training, debugging, or auditing, but disable it during ordinary production inference.
\end{enumerate}

\paragraph{Observation 3.2 (Mathematical non-interference is not engineering zero-cost).}
A1 guarantees that the observation pathway does not alter the mathematical input to the backbone, but the added memory traffic, bandwidth consumption, kernel scheduling, and device synchronization can still reduce throughput. These are separate claims and should not be conflated.

\section{Interpretability: From a Black Box to a Semi-Transparent System}

\subsection{Degree of Transparency}

Bypass Observation does not make a model fully white-box. A more appropriate description is \emph{semi-transparent}, \emph{gray-box}, or \emph{observable gray-box}. Its potential benefits include:

\begin{itemize}
    \item \textbf{(P1) Temporal visibility.}
    Interpretation moves from purely post-hoc analysis toward online monitoring. Layer-wise and token-wise readout may provide an early warning window before a harmful or incorrect conclusion is verbalized in the final output.

    \item \textbf{(P2) Dense coverage.}
    The architecture can, in principle, record every selected layer and every selected token rather than relying only on sparse probe locations.

    \item \textbf{(P3) A read-only observation channel.}
    Because the observation text is not fed back into the backbone, it is not itself part of the model's subsequent causal computation. This makes it qualitatively different from asking a model to retrospectively explain its own answer.

    \item \textbf{(P4) Human-readable outputs.}
    A basic LM-head-style observer exposes vocabulary distributions or top-$k$ token predictions rather than a free-form explanation. A separately trained semantic translator can further compress these signals into natural-language summaries, at the cost of introducing an additional model whose own faithfulness must be validated. Such readouts may also help localize candidate states for later intervention or activation editing~\citep{meng2022,zou2023}.
\end{itemize}

\subsection{Three Fundamental Boundaries}

\begin{itemize}
    \item \textbf{(L1) Lossy verbalization.}
    A vocabulary readout can preserve substantial information about an intermediate hidden state, but converting the full distribution into a small set of tokens or a compact textual summary is necessarily selective. Hidden representations can contain superposed information~\citep{elhage2022}; features that are weakly aligned with vocabulary directions, or that are suppressed by top-$k$ selection or summarization, may be omitted from the human-readable trace.

    \item \textbf{(L2) Correlation is not causation.}
    Observing that a layer appears to ``talk about'' concept $X$ does not imply that concept $X$ is causally responsible for the model's downstream behavior. Activation patching, state perturbation, or related interventions are required to establish a causal link.

    \item \textbf{(L3) The white-box illusion.}
    A well-trained observation head may produce fluent and plausible descriptions even when those descriptions are incomplete or misleading. Readability and faithfulness are independent properties. A fluent misinterpretation can be more dangerous than an obviously noisy one, so deployment requires explicit faithfulness calibration.
\end{itemize}

\paragraph{Observation 4.1.}
Bypass Observation should be interpreted as a potentially faithful \emph{projection} of a computational process, not as a complete description of that process. Attribution claims should therefore be cross-validated with probes, sparse features, and causal interventions.

\subsection{Implications for Safety Auditing}

The main positive effect is increased monitoring density and the possibility of earlier detection. However, the observation pathway also introduces new risks. First, an auditor may confuse fluent translation with actual mechanistic understanding. Second, the observation trajectory itself can become an information-leakage surface that reveals internal states useful for model extraction or attack design. Third, the observation head can itself be corrupted or adversarially manipulated, systematically hiding undesirable signals. Practical systems may therefore need frozen observation heads, integrity checks, access control, output sanitization, and statistical anomaly detection. An asynchronous deployment model---enabled for training and auditing but disabled in ordinary production inference---can reduce some of these risks.

\section{Bypass Chain-of-Thought Versus Conventional Chain-of-Thought}

\subsection{Different Causal Roles}

We call the sequence of layer-wise textual readouts a \emph{bypass chain-of-thought}. Its causal status differs fundamentally from conventional chain-of-thought (CoT), as summarized in Table~\ref{tab:cot}.

\begin{table}[t]
\caption{Causal differences between conventional chain-of-thought and bypass chain-of-thought.}
\label{tab:cot}
\centering
\small
\begin{tabular}{p{0.22\linewidth}p{0.35\linewidth}p{0.35\linewidth}}
\toprule
Dimension & Conventional CoT & Bypass chain-of-thought \\
\midrule
Enters the computation loop
& Yes; generated text becomes part of subsequent context
& No; read-only open-loop observation \\

Causal role of text
& Part of the computation itself
& Projection of hidden computation with no direct causal contribution \\

Generation mechanism
& Autoregressive, serial, context-dependent
& Layer-wise readout parallel to the main generation stream \\

Observation granularity
& Typically one textual trajectory
& Potentially token $\times$ layer observations \\

Text form
& Free-form intermediate reasoning
& Progressively changing predictions or semantic summaries \\

Can provide an RL training signal
& Yes
& Yes; even with a frozen observer, scalar rewards can update the backbone through policy gradients \\

Faithfulness
& May be unfaithful despite being causally active
& Closer to intermediate states, but still lossy and potentially manipulable \\
\bottomrule
\end{tabular}
\end{table}

A useful analogy is that conventional CoT is a draft written by the reasoner, whereas a bypass trajectory is a subtitle generated by an observer. Editing the subtitle does not change the movie. The two trajectories may agree closely, but they may also diverge, especially if the model reaches a decision in middle layers and uses later layers mainly for linguistic refinement.

\subsection{Derived Differences}

Two additional consequences follow:
\begin{itemize}
    \item \textbf{Cross-layer consistency checks.}
    Neighboring observations should often evolve coherently. Abrupt disagreement can itself be a high-information signal, possibly corresponding to representational reorganization or internal conflict.

    \item \textbf{Independent optimization of the observer.}
    Unlike conventional CoT, whose form is coupled to the base model, an observation module can be trained or calibrated independently while the backbone remains frozen.
\end{itemize}

\section{Asymmetric Roles in Reinforcement Learning}

\subsection{Core Distinction}

\paragraph{Observation 6.1 (Asymmetric causal roles).}
Conventional CoT and bypass chain-of-thought occupy different causal positions during inference, but both can provide training signals in reinforcement learning.

Conventional CoT is part of the autoregressive trajectory: once a reasoning token is generated, it enters the subsequent context and directly changes later computation. This gives conventional CoT a dual role. It is both an observable textual trace and a causal component of the inference process. Reinforcement learning from human feedback (RLHF), reinforcement learning with verifiable rewards (RLVR), and process-supervision methods can therefore shape both the generated reasoning trajectory and the downstream computation that consumes it~\citep{ouyang2022,lightman2023,deepseek2025}.

A bypass trajectory is different at inference time because it is read-only: its decoded output is not inserted into the model context and therefore has no direct causal effect on later layers or later tokens. However, this does \emph{not} imply that a frozen bypass observer is unusable for reinforcement learning. Policy-gradient methods do not require the scalar reward to be differentiable with respect to model parameters. A reward computed from a frozen observation head can still update the backbone through the likelihood of sampled trajectories:
\begin{equation}
\nabla_{\theta}J(\theta)
=
\mathbb{E}_{\tau\sim\pi_{\theta}}
\left[
R(\tau)
\sum_t
\nabla_{\theta}\log \pi_{\theta}(a_t\mid s_t)
\right].
\end{equation}
The important distinction is therefore between \emph{inference-time causal participation} and \emph{training-time usefulness as a reward or supervision signal}.

\subsection{Two Training Modes}

Two practically different modes should be distinguished:

\begin{itemize}
    \item \textbf{Mode A: frozen observer.}
    The observation head is fixed and treated as an external measurement function. Its outputs can be used to define scalar rewards, process scores, pseudo-labels, consistency penalties, or data-filtering criteria. The backbone can still be updated through policy gradients or other learning objectives. Freezing the observer improves measurement stability, but it does not eliminate reward hacking: the policy may learn hidden states that score well under the fixed observer without genuinely improving the intended reasoning behavior.

    \item \textbf{Mode B: jointly trained observer.}
    The observer is updated together with the backbone. This increases flexibility and allows the readout to adapt to representation drift, but it also creates an additional failure mode: the observer and policy may co-adapt in a way that improves the measured score without improving the underlying behavior. This is a stronger Goodhart-style risk because both the measured representation and the measurement function can move.
\end{itemize}

\subsection{A Conservative Training Protocol}

A conservative workflow is:

\begin{enumerate}
    \item \textbf{Train or calibrate the observer while freezing the backbone.}
    Possible targets include next-token distributions, layer-specific predictive states, weak supervision from model-generated reasoning traces, or structured semantic labels.

    \item \textbf{Validate the readout before using it as supervision.}
    Use activation patching, state perturbation, controlled counterfactuals, or predictive consistency tests to estimate whether the observation is behaviorally informative. Only sufficiently calibrated layers should be used for high-stakes auditing or reward construction.

    \item \textbf{Prefer auxiliary use before direct optimization.}
    The readout can first be used as an input to a process reward model, a consistency objective, a data-filtering rule, or a pseudo-label generator. These uses reduce the risk of prematurely optimizing directly against a poorly calibrated observer.

    \item \textbf{Avoid circular supervision.}
    The same observation head should not be both the sole generator of a training target and the sole evaluator of whether that target is correct.

    \item \textbf{If direct RL is used, audit for observer gaming.}
    Compare bypass-derived rewards against task success, final-answer correctness, causal interventions, and independently trained observers. A rising bypass score without corresponding behavioral improvement should be treated as a potential reward-hacking signal.

    \item \textbf{Track process-level metrics.}
    Candidate metrics include cross-layer convergence, agreement between intermediate and final predictions, and localization of abrupt state transitions. A reward based on monotonic convergence toward a correct prediction may provide denser process feedback than a purely terminal reward, although this remains an empirical hypothesis.
\end{enumerate}

\section{Implications for Looped Transformers}
\label{sec:looped}

\subsection{The Auditing Problem in Recurrent-Depth Architectures}

Architectures such as the Universal Transformer~\citep{dehghani2019}, Looped Transformers~\citep{giannou2023}, and recurrent-depth latent-reasoning models~\citep{yang2025} reuse weights across multiple computational iterations. This creates several auditing difficulties:

\begin{itemize}
    \item \textbf{(D1) Weight--function decoupling.}
    A fixed statement such as ``layer $k$ performs function $X$'' becomes less meaningful when the same parameters can perform different roles at different iterations.

    \item \textbf{(D2) Input-dependent computational depth.}
    If the number of iterations is adaptive, total computation is not fixed in advance. The process may converge, oscillate, or repeatedly reinforce an incorrect state.

    \item \textbf{(D3) Error localization across iterations.}
    Without intermediate readout, it can be difficult to identify the iteration at which a trajectory begins to deviate.
\end{itemize}

\subsection{Why Bypass Observation Is Particularly Relevant}

\paragraph{Observation 7.1 (From spatial snapshots to a temporal narrative).}
In a standard feed-forward Transformer, layer-wise decoding primarily provides a sequence across depth. In a recurrent-depth model, iteration-wise decoding additionally provides a temporal trajectory. The resulting trace can therefore be interpreted as an observable approximation of the model's convergence process.

\paragraph{Hypothesis 7.2 (Potential halting criterion).}
If an appropriately calibrated observation trajectory becomes stable, that stability may serve as one signal for early stopping. Conversely, persistent oscillation or non-convergence could trigger further computation or an audit alert. This is a hypothesis rather than an established stopping rule and requires receiver operating characteristic (ROC) analysis of false positive and false negative rates.

\paragraph{Observation 7.3 (Per-run iterative auditing).}
Because different inputs may invoke different numbers of iterations, auditing should record the actual trajectory for each run. Bypass Observation naturally provides this per-run window and may help identify infinite loops, oscillations, or self-reinforcing trajectories that are specific to recurrent computation.

\subsection{Additional Costs and Risks}

\begin{itemize}
    \item \textbf{Cost grows with the number of iterations.}
    If a static architecture exposes $L$ observation sites, a recurrent computation repeated for $b$ iterations can create up to $L\times b$ effective observation sites. Sparse or event-triggered monitoring therefore becomes even more important.

    \item \textbf{The observer should be conditioned on time step.}
    The same hidden space can play different functional roles at different iterations. A shared observer that ignores iteration index may therefore be systematically miscalibrated.

    \item \textbf{Apparent convergence can be misleading.}
    Later iterations may show only small output changes even when a small hidden-state update eventually causes a critical decision flip. Differential observation of
    $\Delta\mathbf{h}_t=\mathbf{h}_t-\mathbf{h}_{t-1}$
    may be more informative than static snapshots alone.

    \item \textbf{Feedback is especially risky.}
    Feeding the observation trajectory back into a recurrent computation introduces a new feedback path and may create self-excitation or instability. The read-only constraint is therefore particularly important in looped architectures.
\end{itemize}

\subsection{Unvalidated Engineering Ideas}

Potential engineering mechanisms include gated caching, where only observations containing new information are persisted; ring buffers that retain a recent trajectory; and tiered storage between accelerator memory, CPU memory, and persistent storage. These ideas are currently speculative and require dedicated benchmarking.

\section{Improving the Decoding Rule}

Intermediate-state decoding should not necessarily reuse ordinary greedy language generation. Several alternatives are worth exploring:

\begin{enumerate}
    \item \textbf{Represent distributions rather than only a single token.}
    Intermediate states may encode multiple competing hypotheses. A single greedy token can collapse this uncertainty and create a false sense of confidence. Top-$k$ candidates, entropy, or other distributional summaries may be preferable for faithful inspection.

    \item \textbf{Differential decoding.}
    Decode
    $\Delta\mathbf{h}_{\ell}
    =
    \mathbf{h}_{\ell}-\mathbf{h}_{\ell-1}$
    to ask ``what changed at this layer?'' rather than only ``what does the current state represent?''

    \item \textbf{Cross-layer consistency scoring.}
    A lightweight consistency model can flag abrupt semantic discontinuities between neighboring observations.

    \item \textbf{Observer ensembles.}
    Multiple independently trained observers can be used. Their disagreement should be retained rather than forcibly averaged, because disagreement itself may reveal uncertainty.

    \item \textbf{Structured outputs and capacity constraints.}
    The observer can be restricted to fields such as current subgoal, attended entity, candidate answer, and confidence. Limiting output capacity can reduce the risk that the observer becomes an unconstrained reasoning model with its own hallucinations.

    \item \textbf{Contrastive calibration.}
    Intermediate readouts can be calibrated against later or final predictive distributions to reduce systematic layer-dependent noise.

    \item \textbf{SAE-level captions.}
    Instead of mapping directly into vocabulary space, the observation pathway could expose sparse feature activations and associate them with higher-level semantic descriptions.

    \item \textbf{A dedicated small translator.}
    A separate compact model may be trained to convert hidden states or vocabulary readouts into summaries, providing greater flexibility than directly reusing the base model's LM head. Because this translator performs additional sequence modeling, its runtime cost is not included in the $dV$ closed-form estimate derived in Section~3.
\end{enumerate}

\section{Limitations and Open Questions}

The current proposal has several important limitations:

\begin{enumerate}
    \item \textbf{Faithfulness cannot be self-certified.}
    The usefulness of the entire system depends on whether the observer is faithful, yet faithfulness can only be estimated through external evidence and interventions.

    \item \textbf{White-box illusion.}
    Fluent observation text may lead users to overestimate mechanistic understanding. Structured output, disagreement preservation, and explicit faithfulness scores may help mitigate this problem.

    \item \textbf{Adversarial robustness.}
    A model may learn internal representations that appear benign to an observer while still supporting undesirable computation. The monitoring channel therefore becomes part of the adversarial surface.

    \item \textbf{Practical cost.}
    Full-vocabulary LM-head-style observation at every layer can add roughly 30\%--240\% arithmetic cost under the representative calculations in this paper. A learned semantic translator may incur additional cost beyond this estimate. These costs make always-on deployment unrealistic for many systems.

    \item \textbf{Architectural boundary.}
    Once observation outputs are fed back into the computation, the architecture is no longer Bypass Observation in the strict sense defined by A1. Closed-loop failure modes then reappear.

    \item \textbf{Lack of empirical validation.}
    The present paper is conceptual and analytical. The usefulness of bypass signals for process supervision, the false-alarm rate of recurrent-depth halting criteria, and the information fidelity of low-rank or reduced-vocabulary observers all require systematic experiments.
\end{enumerate}

\section{Conclusion}

This paper develops the concept of Bypass Observation as a non-intrusive, layer-wise semantic readout architecture for large language models.

First, the computational cost of a full-vocabulary readout is primarily governed by the ratio between vocabulary size and effective hidden width. Under a standard approximation, the additional cost is $V/(12d)$, while model-specific calculations for representative architectures range from roughly 30\% to 240\%. Sparse observation, low-rank factorization, reduced vocabularies, and selective decoding can reduce this cost substantially.

Second, Bypass Observation can move a model from a black-box system toward a semi-transparent one by exposing intermediate projections without feeding them back into the backbone. However, such readouts remain lossy, correlational unless validated causally, and vulnerable to the white-box illusion.

Third, bypass chain-of-thought and conventional chain-of-thought occupy fundamentally different causal positions at inference time. Conventional CoT becomes part of the autoregressive context, whereas a bypass trajectory remains read-only. Nevertheless, a frozen bypass observer can still define rewards or supervision signals that update the backbone through policy gradients. This makes bypass readouts suitable for process reward models, pseudo-labeling, consistency checking, auditing, and potentially direct RL, provided that observer gaming is explicitly monitored.

Fourth, recurrent-depth and looped Transformers may be an especially important setting for this architecture. Iteration-wise readout can expose a convergence trajectory, potentially support stopping decisions, and help localize oscillation or self-reinforcing failures. At the same time, recurrent computation amplifies the cost and makes time-step calibration and strict read-only separation more important.

A practical research agenda includes: (1) operational measures of observer faithfulness and intervention-based validation protocols; (2) controlled comparisons of bypass-derived process supervision against outcome rewards and conventional CoT supervision; (3) evaluation of sparse and differential observation for recurrent-depth models, including ROC analysis of halting criteria; (4) access control and anti-probing mechanisms for observation traces; and (5) cross-validation between textual readouts and sparse feature dictionaries.

\subsection*{AI Use Statement}

Generative AI tools were used to assist with language editing, translation, and LaTeX formatting of this manuscript. All technical claims, equations, references, and AI-assisted edits were reviewed by the authors. The authors take responsibility for the final content of the work.

\subsection*{Reproducibility Statement}

This paper is primarily conceptual and analytical. The computational-overhead estimates are derived from the formulas and representative model configurations described in Section~3. The proposed architecture, faithfulness protocols, reinforcement-learning uses, and recurrent-depth applications remain to be validated empirically; no new experimental benchmark is claimed in the present version.

\newpage

\bibliography{iclr2027_conference}
\bibliographystyle{iclr2027_conference}

\end{document}